# Deriving and Combining Continuous Possibility Functions in the Framework of Evidential Reasoning [1]

Pascal Fua

Artificial Intelligence Center, SRI International
333 Ravenswood Avenue
Menlo Park, California 94025

## ABSTRACT

To develop an approach to utilizing continuous statistical information within the Dempster-Shafer framework, we combine methods proposed by Strat and by Shafer. We first derive continuous possibility and mass functions from probability-density functions. Then we propose a rule for combining such evidence that is simpler and more efficiently computed than Dempster's rule. We discuss the relationship between Dempster's rule and our proposed rule for combining evidence over continuous frames.

---

[1] The work reported here was supported in part by the Defense Advanced Research Projects Agency under Contract MDA903-83-C-0027 and in part by the Ecole Nationale Supérieure des Télécommunications.

# 1  Introduction

The knowledge used in many expert systems can often be expressed in the form of conditional probabilities, typically the conditional probability of making a particular observation when a given state holds. The Dempster-Shafer paradigm provides a framework for dealing with nonstatistical information, representing ignorance and combining evidence coming from multiple, independent knowledge sources. We propose an approach to the representation of statistical information within this framework, in order to combine it with information from other sources.

While Shafer [1] shows how this goal can be achieved for discrete probability distributions, probability distributions of interest are often continuous. Using Strat's technique [2], we can derive mass functions from individual probability-density functions. To combine evidence we can use Dempster's rule; however, in this context, Dempster's rule is computationally expensive [3] and yields very complex mass functions without a simple intuitive interpretation. We therefore propose a rule for evidence combination that yields simpler results and show that this rule is both statistically sound and related to Dempster's rule.

We begin by explaining how possibility and mass functions can be derived from a probability-density function. We then prove that the evidence can be pooled simply in a way that is statistically sound and amounts to an approximation of Dempster's rule. Finally we introduce an example showing how this result can be generalized to combine evidence that is statistical in nature with evidence that is not.

# 2  Mass Function From Probability Density Functions

In this section, we show how the basic method for statistical inference described by Shafer [1] can be used to handle probability-density functions.

Suppose that an expert has stated his knowledge about a system in the form of the conditional probability of making an observation $f$ when the system is in state $\theta$, for all $f$ and $\theta$. Let $F$ be the set of possible observations and $\Theta$ the set of possible states. Suppose that we now perform an experiment and observe $f$. Our goal is to assess our beliefs about the possible states of the system.

If the prior probabilities of the states are available, Bayes' rule allows us to achieve this goal very simply. However, if they are not, beliefs can be assessed in the following way:

We shall assume that the evidence favors the states $\theta \in \Theta$ that maximize the probability of observing the actual outcome of the experience, which amounts to a maximum-likelihood estimate. We therefore define a *possibility function*, *poss* that plays the same role as the possibility function in Zadeh's theory [5] and is such that

$$\forall \theta \in \Theta, \ poss(\theta) \propto p(f|\theta) \ .$$

If our evidence is unbiased and is free of systematic errors, there is no reason to disbelieve the results of the experiment. At least one state must be completely possible: we normalize *poss* so that

$$\max_\theta poss(\theta) = 1 \ .$$

Such evidence can be represented by a consonant mass function *mass* and associated plausibility function $pl$ such that

$$\forall \theta \in \Theta, \ pl(\{\theta\}) = poss(\theta) \ .$$

In this work, we restrict ourselves to the case in which the evidence bears upon scalar quantities and points in a single direction, but with some uncertainty. $\Theta$ can then be considered as an interval in the set of real numbers, and we shall assume that the possibility functions are unimodal, strictly monotonically decreasing about their maximum, differentiable, and have zero value on the boundary of $\Theta$.

As shown by Strat [2], under the above assumption such a consonant mass function exists and is unique. Because *poss* is unimodal and strictly monotonically decreasing about $x_m$, there exists a function $u : [A, x_m] \mapsto [x_m, B]$ such that $\forall l \in [A, x_m]$ $poss(l) = poss(u(l))$. Then, *mass* is defined as the mass function with focal elements the intervals $[l, u(l)]$ such that

$$mass([l, u(l)]) = \frac{\frac{d[poss(l)]}{dl}}{(1 + \frac{du(l)}{dl}^2)^{\frac{1}{2}}} \quad . \quad (1)$$

## 3 Combining the Evidence

For these continuous possibility and mass functions to be useful, we need to be able to combine the evidence from several knowledge sources.

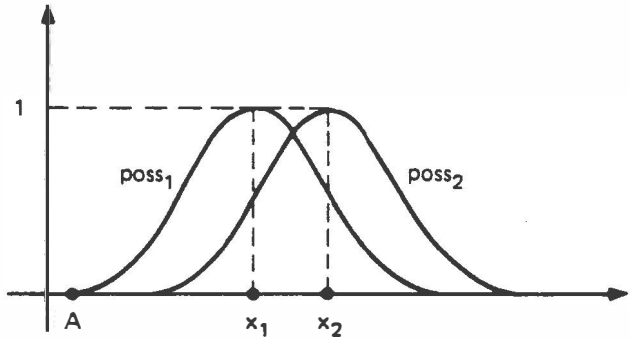

Figure 1: Two Possibility Functions

We first define the following notation that will be used in the rest of this paper: Suppose that we perform two independent measurements $f_1$ and $f_2$. From them, we can derive two possibility functions $poss_1$ and $poss_2$, shown in Figure 1, with maxima $x_1$ and $x_2$ and two corresponding mass functions $m_1$ and $m_2$ over $\Theta = [A, B]$.

### 3.1 A Simple Rule

The possibility functions $poss_1(\theta)$ and $poss_2(\theta)$ are proportional to the conditional probabilities $p_1(f_1|\theta)$ and $p_2(f_2|\theta)$. If the two measurements are independent, then

$$\forall \theta \in \Theta, \; p(f_1, f_2|\theta) = p_1(f_1|\theta) p_2(f_2|\theta) \quad .$$

Therefore a simple and statistically sound way to combine the evidence is to *multiply the possibility functions pointwise, normalize the product and use it to derive a new consonant mass function*, as described above (See Shafer [4] for a comparison of alternative techniques for combining possibility functions).

### 3.2 Comparison with Dempster's rule

Another way to combine evidence is to use Dempster's rule. But when we combine our continuous mass functions, which have an infinite set of focal elements, a severe problem arises from the fact that the mass function resulting from the combination will be extremely complex and will not possess any simple intuitive meaning. In particular, it will be highly nonconsonant.

Let $m_{poss}$ be the mass function we derive from the product of the possibility functions by assuming consonance, and let $m_{Dempster}$ be the mass function that we obtain by "dempsterizing" $m_1$ and $m_2$. Let $pl_{poss}$ and $pl_{Dempster}$ be the corresponding plausibility functions.

In general, $mass_{Dempster}$ will be nonconsonant and different from $mass_{poss}$. But the two mass functions assign proportional masses to singletons:

$$\forall \theta \in \Theta$$
$$pl_{Dempster} = \frac{1}{1-k} pl_1(\{\theta\}) pl_2(\{\theta\})$$
$$pl_{poss} = \frac{1}{norm} pl_1(\{\theta\}) pl_2(\{\theta\}) \quad ,$$

where $k$ is Dempster's contradiction factor and $norm = max_{x \in \Theta}(poss_1(x) poss_2(x))$.

Let $a = 1 - k$ and $x_0$ be the point at which $poss_1(x) poss_2(x)$ reaches its maximum. Since $\forall x \in \Theta, \; pl_{Dempster}(\{x\}) \leq 1$,

$$pl_{Dempster}(\{x_0\}) = \frac{norm}{a} \leq 1 \quad .$$

Therefore $a$ is always greater than $norm$ and

$$\forall \theta \in \Theta, \; pl_{Dempster}(\{\theta\}) \leq pl_{poss}(\{\theta\}) \quad .$$

This observation is the key to understanding the relationship between our multiplicative combination rule and Dempster's rule. Plausibilities can be viewed as upper bounds on probabilities; therefore, $pl_{poss}$ is less informative than $pl_{Dempster}$. But, because these bounds are consistently bigger, no inconsistency with respect to Dempster's rule has been introduced. This is intuitively appealing because, in order to derive $m_{poss}$, we have assumed consonance. We have "cancelled out" the nonconsonance of $m_{Dempster}$ that resulted from the conflict between $m_1$ and $m_2$, thereby losing some information. In particular, if the two possibility fuctions have the same maximum, there is no conflict at all between them and it can be shown that the two combination rules yield exactly the same result.

We can thus combine the evidence by simply multiplying the possibility functions pointwise. We lose some information about the conflict between the two knowledge sources, but we shall show in the following subsection that the magnitude of that conflict can be estimated easily in this context.

### 3.3 Agreement Between Two Knowledge Sources

In the Dempster-Shafer framework, the conflict between two knowledge sources is represented by Dempster's contradiction factor $k$.

For discrete mass functions $m_1$ and $m_2$, with corresponding plausibility functions $pl_1$ and $pl_2$, $k$ is defined by:

$$k = \sum_{F,G \subset \Theta \text{ and } F \cap G = \emptyset} m_1(F) m_2(G) \ .$$

Therefore, if we call $a = 1 - k$ the *agreement* between the two mass distributions, we can show that

$$a = \sum_{F \subset \Theta} m_2(F) pl_1(F) = \sum_{F \subset \Theta} m_1(F) pl_2(F) \ .$$

We now replace the discrete mass functions with the continuous ones defined above. If we assume that $x_1 \leq x_2$, and replace $m$ and $pl$ with their expression in terms of *poss* as in Eq. 1, the sum becomes an integral and we find the formula

$$\begin{aligned} a &= poss_2(x_1) + \int_{x_1}^{x_2} poss_2'(l) poss_1(l) dl \\ &= poss_1(x_2) - \int_{x_1}^{x_2} poss_1'(l) poss_2(l) dl \ , \end{aligned}$$

where $poss'(x)$ is the derivative of *poss* with respect to its argument.

The agreement between two knowledge sources can therefore be computed directly from the possibility functions *without computing the mass functions*.

In summary, by using our simple multiplicative combination rule, we do not record the conflict between the two knowledge sources and therefore lose some information. But, because we can still evaluate Dempster's contradiction factor without computing the actual mass functions, we are able to ascertain how good the evidence is.

## 4 Example

So far we have seen that we can derive continuous possibility and mass functions from probability-density functions and combine them in a statistically consistent manner. In the following example, we show how we can extend the scope of these ideas to combine statistical information with nonstatistical information that has an obvious representation in the Demspster-Shafer formalism.

Let us suppose that we are in court trying to determine whether a car was speeding or not. We have two pieces of evidence:

1. A police officer has clocked the car with his radar, the characteristics of which are known. The technique described above can be used to represent this information by a possibility function $poss_r$ peaking at $v_r$ shown in Figure 2(a).

2. A witness has testified that he was following the car and that, according to his speedometer, its speed remained betwen $v_1$ and $v_2$.

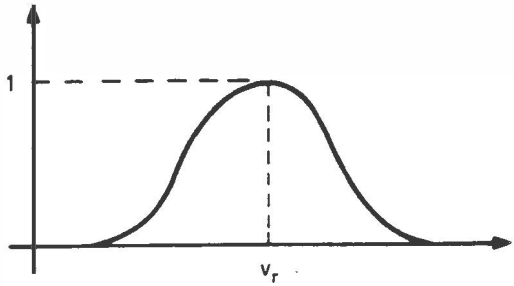

(a) Radar report

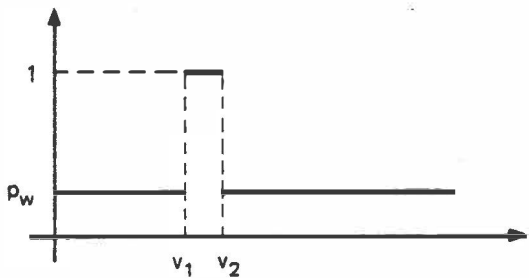

(b) Witness report

Figure 2: Representing the evidence using possibility functions.

Suppose we can assess the validity of such testimony and that there is a probability $1-p_w$ for its being accurate. In other words, with probability $p_w$ the witness is unreliable and his testimony is not informative. If the inherent inaccuracy of the speedometer is small compared with the width of the interval $[v_1, v_2]$, this information can be naturally represented as a simple support function, with focal elements $[v_1, v_2]$ and $\Theta$ with respective masses $1 - p_w$ and $p_w$. So far we have derived mass functions from possibility functions, conversely, in this case we can define a possibility function $poss_w$ as the contour function of our simple support function, as shown in Figure 2(b).

Multiplying the possibility functions pointwise can be viewed as an approximation of Dempster's rule; we therefore combine the evidence by computing and normalizing the product of the possibility functions, as shown in Figure 3. Although $poss_w$ is no longer continuous, the agreement $a$ can still be computed simply, yielding

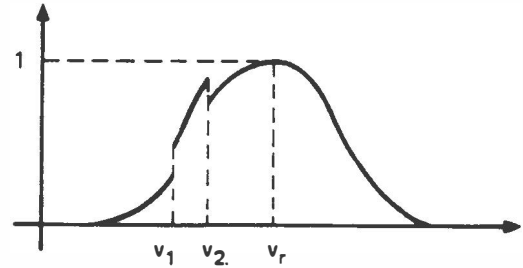

Figure 3: Combining the two reports.

$$a = (1 - p_w)poss_r(v_2) + p_w$$
$$norm = max(poss_w(v_2), p_w)$$

If $v_1 \leq v_r \leq v_2$, the two reports agree, $a = 1$ and there is no evidence against the maximum-likelihood estimate $v_r$. If $v_2 < v_r$, the two reports start conflicting and the smaller $v_2$ is, the greater the conflict. The maximum-likelihood estimate is either $v_2$ or $v_r$, depending on which report is more reliable; it can be shown that $1 - \frac{norm}{a}$ represents the support, in the Dempster-Shafer sense, of the proposition "the speed is different from the maximum-likelihood estimate" and can be used as a measure of the validity of this estimate.

By multiplying the two possibility functions, we are able to recover useful information for decision-making. We avoid computing and "dempsterizing" the actual mass functions, which could be computationally expensive. Furthermore, the result is a new possibility function that can again be combined in the same fashion with additional evidence.

## 5 Conclusion

We have shown that in the case of continuous scalar quantities, we can use the Dempster-Shafer approach to derive possibility functions and beliefs from probability distributions. The evidence can then be pooled very easily by multiplying the possibility functions pointwise. The approach is a consistent approximation of Dempster's rule, and permits one to estimate simply the conflict between knowledge sources. In future work, we shall generalize this approach to multivariate distributions. Because beliefs are expressed in the framework of evidential reasoning, they can be combined with those generated by any other kind of knowledge source; this is a very desirable feature for automated decision-making.